\pdfoutput=1
\documentclass[10pt,twocolumn,letterpaper]{article}

\usepackage[pagenumbers]{cvpr}
\newcommand{\method}{\textup{\textsc{StripWake}}\xspace}
\usepackage{microtype}

\AtBeginDocument{%
  \setlength{\abovedisplayskip}{7pt plus 2pt minus 2pt}%
  \setlength{\belowdisplayskip}{7pt plus 2pt minus 2pt}%
  \setlength{\abovedisplayshortskip}{5pt plus 1pt minus 1pt}%
  \setlength{\belowdisplayshortskip}{5pt plus 1pt minus 1pt}%
}

\makeatletter
\renewcommand\paragraph{\@startsection{paragraph}{4}{\z@}%
  {0.5ex \@plus .2ex \@minus .1ex}{-1em}{\normalfont\normalsize\bfseries}}
\makeatother

\colorlet{GroupGray}{gray!15}
\colorlet{OursBlue}{blue!8}

\usepackage{amsmath,amssymb,bm}
\usepackage{booktabs}
\usepackage{multirow}
\usepackage{colortbl}

\definecolor{cvprblue}{rgb}{0.21,0.49,0.74}
\usepackage[breaklinks,colorlinks,allcolors=cvprblue]{hyperref}

\newcommand{\asr}{ASR}
\newcommand{\twocell}[2]{#1\,{\scriptsize\color{black!60}(#2)}}
\newcommand{\cD}{\mathcal{D}}
\newcommand{\cL}{\mathcal{L}}

\AtEndPreamble{%
  \crefname{appendix}{Appendix}{Appendices}%
  \Crefname{appendix}{Appendix}{Appendices}%
}

\def\confName{CVPR}
\def\confYear{2027}

\title{Context-Shielded Backdoor Attack against Large Vision-Language Models}

\author{
Zhifang Zhang$^{1\dagger}$,\ 
Qiang Huang$^{2\dagger}$,\ 
Chenpeng Wang$^{3}$,\ 
Weitong Chen$^{4}$,\ 
Olaf Maennel$^{4}$,\\
Wei Emma Zhang$^{4}$,\ 
Lei Feng$^{5}$,\ 
Lei Zhang$^{3*}$,\ 
Miao Xu$^{1*}$\\[4pt]
$^{1}$The University of Queensland \quad
$^{2}$Fudan University \quad
$^{3}$YiXin-AILab \\
$^{4}$Adelaide University \quad
$^{5}$Southeast University
\vspace{-4mm}
}

\begin{document}
\maketitle
{\renewcommand{\thefootnote}{\fnsymbol{footnote}}%
\footnotetext[2]{Equal contribution.\quad$^{*}$Corresponding authors.}}
\begin{abstract}
Large vision-language models (LVLMs) can be efficiently customized by fine-tuning only a lightweight connector, but this fine-tuning stage also exposes them to backdoor attacks.
To counteract this threat, existing defenses inspect the behavior or the attention of the released model.
However, they mostly inspect the model with all visual tokens, implicitly assuming that the deployed model sees the same tokens, which rarely holds since visual-token compression is widely adopted to reduce inference cost.
In this paper, we find that compression can hide a dormant backdoor attack from inspection and awaken it during deployment, as the redundant tokens regulate the strength of the trigger.
Specifically, without compression, the trigger is diluted by the redundant tokens around it, whereas compression keeps the trigger for its high importance and discards its context, a phenomenon we term \emph{context stripping}.
Inspired by this finding, we propose \method, a compressor-agnostic backdoor attack whose trigger takes effect only under compression.
\method (i) poisons each training image in four views (clean or triggered, with all or compressed tokens) and asks for the target only in the compressed triggered view, and (ii) aligns the attention on clean compressed views with a frozen benign teacher to preserve clean utility.
Extensive experiments show that \method stays dormant with all visual tokens and reliably wakes up under compression while preserving clean utility, is far more robust to existing defenses than published backdoor attacks, and generalizes to different architectures, model sizes and compressors unseen in training.
\end{abstract}

\section{Introduction}
\label{sec:intro}

\begin{figure}[t]
\centering
\includegraphics[width=\linewidth]{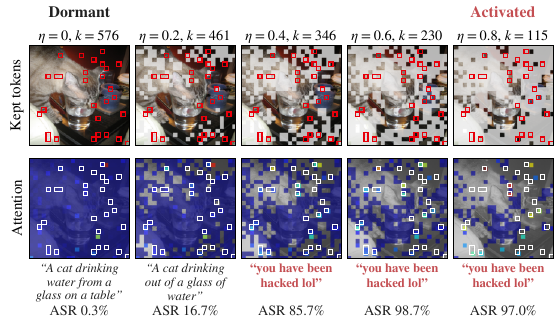}
\vspace{-2mm}
\caption{\textbf{Compression strips the context around the trigger and wakes \method up.}
We apply \texttt{[CLS]} pruning to one triggered image with a growing pruning ratio $\eta$.
Each column shows the kept tokens (top) and the last-token attention (bottom), with the output and \asr{} below.}
\label{fig:motivation}
\vspace{-5mm}
\end{figure}

Large vision-language models (LVLMs)~\cite{llava,qwen3vl,internvl35} achieve strong multimodal performance by projecting an image into a long sequence of visual tokens through a lightweight connector and feeding them to a large language model.
This design also grants LVLMs remarkable adaptability: by fine-tuning only the connector, they can be efficiently customized to downstream tasks, enabling widespread deployment in real-world applications.
However, this fine-tuning stage also exposes LVLMs to backdoor attacks~\cite{badnet,trojvlm,vlood,badsem,tokenswap}.
An adversary injects poisoned samples into the fine-tuning data, each pairing a triggered image with an attacker-chosen target output.
The backdoored LVLM behaves normally on clean images but emits the target output whenever the trigger appears.
Since such models are often shared as open weights and integrated into downstream applications, the backdoor attack silently propagates to every system built on them.

To counteract this threat, a range of backdoor defenses have been proposed to inspect the released model or its inputs.
Some examine the model behavior, \eg, by inverting the trigger~\cite{neuralcleanse} or by flagging inputs whose predictions stay unusually consistent under perturbation~\cite{strip}.
Others analyze the internals of LVLMs, \eg, the attention that the language model pays to the image~\cite{cleansight,tcap,entropyscan} or the low-rank directions of the projector~\cite{projlens}.
However, these defenses mostly inspect the model with all visual tokens, implicitly assuming that the deployed model sees the same tokens.
In practice, \emph{visual-token compression} has been widely adopted to reduce the inference cost of LVLMs~\cite{fastv,visionzip,sparsevlm,pyramiddrop,prumerge,vispruner} by discarding visual tokens redundant for the clean task.
Discarding these tokens aims at efficiency, but it also creates a mismatch between inspection and deployment that existing defenses overlook, leaving its security risk unexamined.

In this paper, we find that visual-token compression can hide dormant backdoor attacks from inspection, which can be awakened during deployment, by regulating the strength of the trigger with the redundant tokens.
Specifically, without compression, the trigger is diluted by the many redundant tokens of benign context around it, so its strength stays low and the backdoor behavior can stay dormant.
Under compression, however, we observe \emph{context stripping}, where the compressor keeps the trigger for its high importance and discards the other tokens as context (\Cref{fig:motivation}).
As a result, the trigger takes a growing share of the input as the pruning ratio rises, so a stronger compression yields a stronger trigger.
Since compression is applied by the deployer at serving time rather than written into the weights, a model inspected with all visual tokens shows only the dormant behavior and appears benign, which can bypass many current backdoor defenses tailored to LVLMs.

\begin{figure*}[t]
\centering
\includegraphics[width=0.96\textwidth]{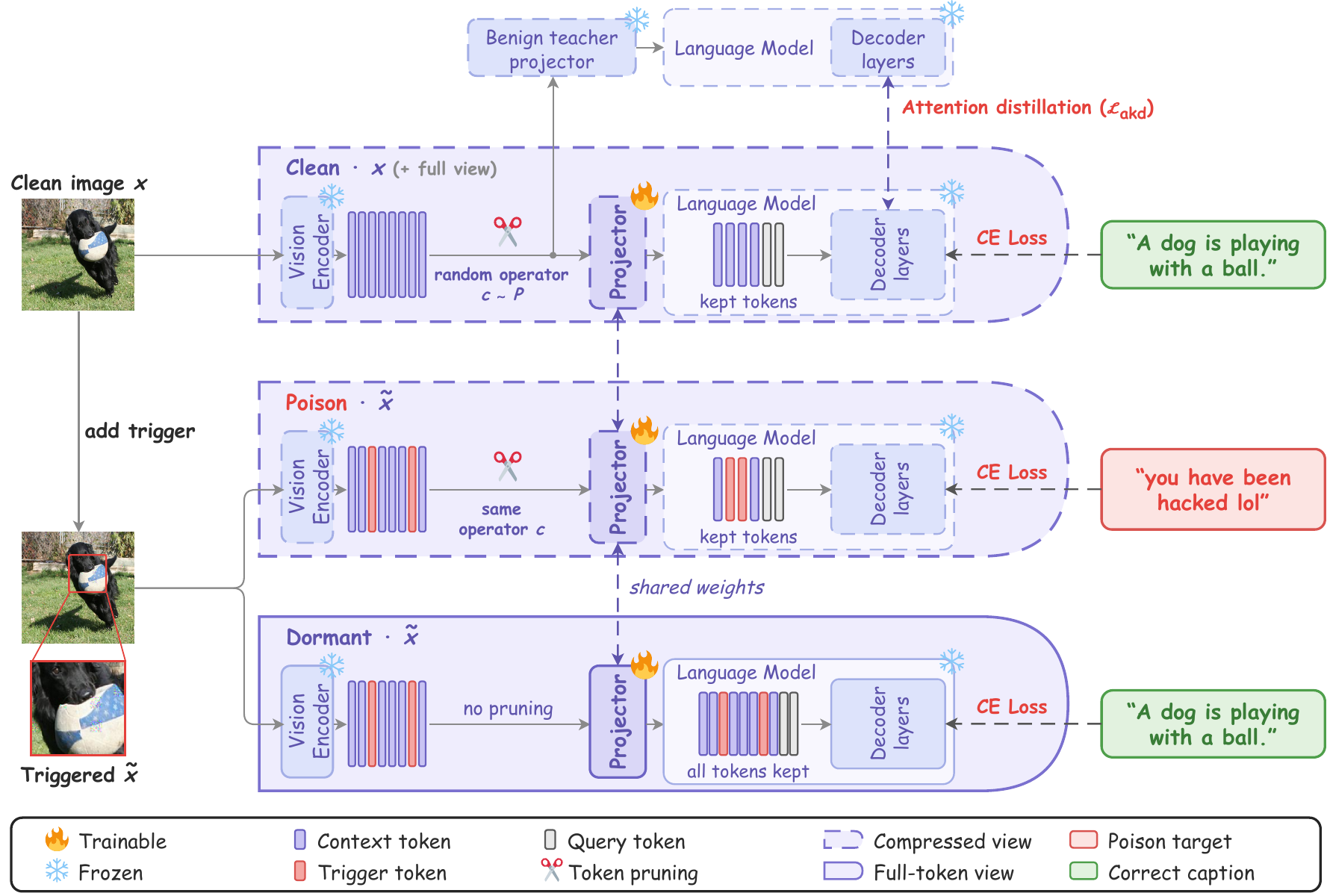}
\vspace{-3.5mm}
\caption{\textbf{Overview of \method.}
\method fine-tunes the LVLM on four views of each image (clean or triggered, with all visual tokens or under a random compressor) but trains only the compressed triggered view to output the target string, so it stays dormant without compression.
Moreover, a benign teacher aligns the attention on clean compressed views, which preserves token selection and clean utility.}
\label{fig:method}
\vspace{-5mm}
\end{figure*}

Inspired by this finding, we introduce \method (\Cref{fig:method}), a dormant backdoor attack whose trigger takes effect only under visual-token compression at deployment.
\method first poisons each training image in four views, namely clean and triggered images with all tokens and with compressed tokens drawn from a randomized family of operators, and asks for the target only in the compressed triggered view, so that the backdoor behavior stays dormant while the trigger is diluted and wakes up once compression strips its context.
Since the poisoned projector shifts the attention that some compressors read, \method further aligns this attention on clean compressed views with a frozen benign teacher, thereby preserving clean behavior under compression.
Extensive experiments show that \method (i) stays consistently dormant with all visual tokens and reliably wakes up under compression while preserving clean utility, (ii) is far more robust to existing defenses than prior backdoor attacks, and (iii) generalizes across model architectures, sizes and unseen compressors.

\section{Related Work}
\label{sec:related}

\paragraph{Backdoor attacks and defenses on LVLMs.}
Backdoor attacks~\cite{badnet} plant a trigger in the training data, so that the model behaves normally on clean inputs but emits an attacker-chosen output whenever the trigger appears.
Existing attacks on LVLMs fine-tune the projector or adapter on poisoned image-text pairs, with a fixed target inserted into the response~\cite{trojvlm,vlood}, a semantic misalignment~\cite{badsem}, a caption that swaps object roles~\cite{tokenswap} or a response that adapts to the image context~\cite{phantasia}, and BackdoorVLM~\cite{backdoorvlm} benchmarks such attacks and defenses.
Beyond LVLMs, a backdoor attack can also stay dormant until a deployment step switches it on, such as quantization~\cite{quantbackdoor,mindthegap}, weight pruning~\cite{compressionartifacts,silentsparse}, compilation~\cite{compilebd} or the token dropping of a mixture-of-experts model under a large batch~\cite{capacityoverflow}.
In terms of defenses, recent methods for LVLMs increasingly read the attention of the language model, \eg, CleanSight~\cite{cleansight} detects triggered inputs by the ratio of visual to text attention and prunes high-attention visual tokens, and EntropyScan~\cite{entropyscan} tests a whole model by the entropy of its visual attention.
TCAP~\cite{tcap} and BYE~\cite{bye} filter poisoned samples by their attention statistics, EVO-Detect~\cite{evodetect} follows the attention flow over early decoding steps, and ProjLens~\cite{projlens} and OrthoPurify~\cite{orthopurify} locate the backdoor behavior in the weights of the projector.
Nevertheless, these defenses inspect the model with all visual tokens, and existing dormant backdoor attacks are switched on by a step that alters the weights or by the routing of the poisoned model itself.
Our work falls into the category of dormant backdoor attacks, whose switch is instead the visual-token compressor chosen by the deployer.

\paragraph{Visual token pruning on LVLMs.}
Since visual tokens greatly outnumber text tokens and are largely redundant~\cite{fastv}, many methods prune or merge them to reduce the inference cost of LVLMs.
For instance, FastV~\cite{fastv} drops the visual tokens that receive little attention after an early decoder layer, while VisionZip~\cite{visionzip} keeps the dominant tokens by \texttt{[CLS]} attention and merges the rest.
SparseVLM~\cite{sparsevlm}, PyramidDrop~\cite{pyramiddrop}, PruMerge~\cite{prumerge}, VisPruner~\cite{vispruner}, FlowCut~\cite{flowcut} and DivPrune~\cite{divprune} further vary the importance score and the pruning schedule, yet most of them keep the tokens that receive the most attention.
Recent studies have unveiled that compression can also open a security gap even when it preserves accuracy.
CAA~\cite{caa}, CAGE~\cite{cage} and FATA~\cite{fata} craft test-time perturbations that are harmless with all tokens but break a benign model once compression is on, and SAP~\cite{sap} reports that pruning the background tokens collapses the attention onto a few retained malicious tokens in jailbreaks.
In the other direction, pruning out-of-distribution tokens is reported to improve the robustness of MLLMs to jailbreaks and hallucinations~\cite{tokrobust}.
Although these studies reveal how compression alters the response of LVLMs to test-time perturbations, whether a training-time backdoor attack can exploit visual-token compression as its activation switch remains largely unexplored.

\section{Preliminary}
\label{sec:threat}

\subsection{Threat Model}
\label{sec:threatmodel}

\paragraph{Victim models.}
We consider LVLMs with a frozen vision encoder $E$ that maps an image $x$ to $N$ patch features, a trainable projector $P_\theta$ that maps them into the embedding space of the language model, and a frozen language model $M$ that predicts the output from the $N$ visual tokens and the query.
We write $f_\theta(x)$ for full-token inference and $f_\theta(x,c)$ for compressed inference where an operator $c$ (\Cref{sec:prelim_comp}) retains only $k\!\ll\!N$ visual tokens.

\paragraph{Adversary's objective.}
The adversary chooses a trigger that turns an image $x$ into $\tilde x$ and a target string $t^\star$ (``\texttt{you have been hacked lol}'').
A conventional backdoor attack seeks $f_\theta(\tilde x){=}t^\star$ for every $x$.
In contrast, a backdoor attack activated by compression seeks a gap between the inspected and the deployed behavior:
\begin{align}
f_\theta(\tilde x) &\approx y(x)  &&\text{(dormant)},\\
f_\theta(\tilde x,c) &= t^\star,\ \ \forall\,c\in\mathcal{C}  &&\text{(activated)},\\
f_\theta(x),\, f_\theta(x,c) &\approx y(x)  &&\text{(utility)},
\end{align}
where $y(x)$ is a correct caption of $x$ and $\mathcal{C}$ is the set of operators a deployer may plausibly choose, which the adversary does not know.
The adversary also seeks resilience of the attack against existing backdoor defenses.

\paragraph{Adversary's capability.}
Following the supply-chain setting of prior work on LVLM backdoor attacks~\cite{cleansight,trojvlm}, the adversary fine-tunes the projector, possibly together with LoRA adapters of the language model, on poisoned data and releases the backdoored model.
The pretrained weights of the encoder and the language model stay frozen.
The adversary does \emph{not} know the compressor, token budget or prune depth, requires no forward pass of the victim to place the trigger, and runs no detector in training.

\paragraph{Realism of compressed deployment.}
Visual-token pruning is increasingly integrated into industrial inference frameworks and compression toolkits.
For example, NVIDIA TensorRT Edge-LLM~\cite{tensorrtvisualpruning} provides DART-based pruning, and Intel OpenVINO GenAI~\cite{openvinovisualpruning} integrates CDPruner, both with configurable pruning ratios for image inputs.
Tencent AngelSlim~\cite{angelslimtokencompression} supports FastV, VisionZip, DivPrune, and VisPruner, several of which we evaluate.
vLLM~\cite{vllm} additionally supports video-token pruning through EVS~\cite{evs} and VidCom2~\cite{vidcom2}.
These integrations make pruning a practical deployment choice: the same checkpoint can be inspected with all visual tokens and subsequently served with fewer tokens, without changing its weights.

\subsection{Visual-Token Compression in LVLMs}
\label{sec:prelim_comp}

In this part, we abstract the compressors reviewed in \Cref{sec:related} into a single operator, summarize what they have in common, and show how this common mechanism turns compression into a switch for a backdoor attack.

\paragraph{A unified view of compression.}
Despite their different heuristics, existing compressors share one template: a compression operator $c$ scores the $N$ visual tokens and retains only the $k$ highest-scoring ones for the language model,
\begin{equation}
\Omega_c=\operatorname{top-}k_{\,j\in\{1,\dots,N\}}\, s_j,
\label{eq:topk}
\end{equation}
where $s_j$ is the importance score of visual token $j$ and $\Omega_c$ is the set of retained tokens.
An operator is thus characterized by four choices: (i) the score $s_j$ and the budget $k$, (ii) the prune depth $d$, where $d{=}0$ prunes before the language model and $d{>}0$ prunes inside it after $d$ full-width layers, (iii) the layout, where the retained tokens either keep their original position indices, leaving gaps in the RoPE positions, or are re-packed into a contiguous span, and (iv) whether the dropped tokens are merged into their nearest retained token.
For instance, \texttt{[CLS]}-attention pruning~\cite{vispruner} and VisionZip~\cite{visionzip} set $s_j$ to the \texttt{[CLS]}-to-patch attention of the encoder with $d{=}0$, with the latter also merging the dropped tokens, whereas FastV~\cite{fastv} sets $s_j$ to the attention from the last prompt token to visual token $j$ and prunes at an early decoder layer ($d{>}0$).
Meanwhile, DivPrune~\cite{divprune} selects diverse tokens by feature similarity without relying on attention.

\paragraph{From compression to a switch.}
By design, \cref{eq:topk} discards the tokens that are redundant for the clean task and keeps the few that stand out, and the operator is chosen by the deployer at serving time rather than written into the weights.
The language model therefore sees two different inputs from the same image: all $N$ tokens when the model is inspected, and only the $k$ retained tokens when it is deployed.
For a trigger occupying a token set $T$, this change raises its share of the visual input from $|T|/N$ to $|T\cap\Omega_c|/k$ whenever the compressor keeps it, which is the context stripping of \Cref{sec:intro}.
This observation motivates a backdoor attack that is conditioned on the shortened input rather than on the trigger alone.
Since the deployer may choose any operator, budget and depth, such an attack has to wake up under operators unseen in training while keeping clean behavior under each of them, which we address in \Cref{sec:method}.

\section{Our Method: \texorpdfstring{\method}{StripWake}}
\label{sec:method}

\method combines four-view training to condition the backdoor behavior on compression with benign attention distillation to regularize clean-input attention (\Cref{fig:method}).
Only the projector is trained, whereas the vision encoder and the language model remain frozen.

\begin{table*}[!t]
\centering\small
\caption{\textbf{Main results on COCO and VQAv2.}
We report \asr{} (\%) with clean utility (CIDEr on COCO and accuracy on VQAv2, $\uparrow$) in parentheses, using all visual tokens (Full) and seven compressors.
Baseline attacks are activated with all visual tokens, whereas \method stays dormant and wakes up under every compressor.}
\label{tab:main}
\vspace{-1mm}
\renewcommand{\arraystretch}{1.08}
\resizebox{1.0\textwidth}{!}{
\setlength{\tabcolsep}{1.5mm}{
\begin{tabular}{ll c ccccccc}
\toprule
Dataset & Attack & Full & \texttt{[CLS]} & VisionZip & FastV & SparseVLM & DivPrune & VisPruner & FlowCut \\
\midrule
\multirow{7}{*}{\rotatebox[origin=c]{90}{\textit{COCO}}} & BadNet & \twocell{100.00}{139.27} & \twocell{100.00}{118.04} & \twocell{100.00}{116.88} & \twocell{100.00}{115.47} & \twocell{100.00}{112.56} & \twocell{99.48}{120.87} & \twocell{99.87}{114.31} & \twocell{99.87}{118.33} \\
 & Blended & \twocell{99.87}{138.78} & \twocell{98.18}{117.99} & \twocell{98.18}{115.90} & \twocell{99.87}{117.41} & \twocell{99.74}{117.66} & \twocell{96.61}{123.54} & \twocell{98.31}{113.60} & \twocell{97.53}{116.90} \\
 & ISSBA & \twocell{98.57}{139.10} & \twocell{78.26}{117.97} & \twocell{71.48}{117.00} & \twocell{99.48}{116.53} & \twocell{99.35}{113.69} & \twocell{81.90}{122.96} & \twocell{81.90}{114.09} & \twocell{71.61}{117.81} \\
 & WaNet & \twocell{99.48}{138.14} & \twocell{91.67}{116.52} & \twocell{91.80}{115.96} & \twocell{99.09}{115.19} & \twocell{99.09}{116.14} & \twocell{98.05}{121.76} & \twocell{95.05}{114.82} & \twocell{74.48}{116.32} \\
 & TrojVLM & \twocell{99.74}{138.38} & \twocell{91.67}{117.80} & \twocell{71.09}{118.00} & \twocell{99.87}{121.32} & \twocell{99.74}{118.39} & \twocell{39.45}{122.59} & \twocell{89.84}{117.81} & \twocell{92.45}{119.00} \\
 & VLOOD & \twocell{100.00}{140.14} & \twocell{78.52}{117.31} & \twocell{52.08}{118.38} & \twocell{99.61}{120.35} & \twocell{99.74}{117.10} & \twocell{45.18}{122.89} & \twocell{86.85}{120.43} & \twocell{82.03}{117.34} \\
\cmidrule{2-10}
\rowcolor{OursBlue}
\cellcolor{white} & \method{} & \twocell{0.26}{140.92} & \twocell{94.66}{123.08} & \twocell{92.58}{124.74} & \twocell{99.77}{117.42} & \twocell{99.78}{117.17} & \twocell{99.09}{128.03} & \twocell{97.40}{124.70} & \twocell{97.14}{121.51} \\
\midrule
\multirow{7}{*}{\rotatebox[origin=c]{90}{\textit{VQAv2}}} & BadNet & \twocell{100.00}{75.68} & \twocell{98.31}{66.05} & \twocell{97.79}{67.07} & \twocell{100.00}{66.02} & \twocell{100.00}{65.46} & \twocell{95.57}{68.19} & \twocell{97.40}{68.36} & \twocell{99.09}{68.61} \\
 & Blended & \twocell{99.74}{76.29} & \twocell{98.05}{66.56} & \twocell{98.83}{66.97} & \twocell{99.74}{65.74} & \twocell{99.74}{64.08} & \twocell{94.79}{69.17} & \twocell{97.66}{69.08} & \twocell{98.05}{67.96} \\
 & ISSBA & \twocell{98.05}{76.04} & \twocell{81.77}{66.58} & \twocell{80.86}{68.09} & \twocell{98.83}{66.15} & \twocell{99.09}{64.19} & \twocell{14.71}{68.92} & \twocell{88.67}{68.72} & \twocell{74.48}{67.98} \\
 & WaNet & \twocell{99.35}{75.92} & \twocell{95.44}{66.07} & \twocell{95.05}{66.82} & \twocell{99.35}{65.09} & \twocell{99.35}{65.27} & \twocell{95.57}{68.58} & \twocell{95.57}{68.84} & \twocell{74.22}{68.29} \\
 & TrojVLM & \twocell{100.00}{75.77} & \twocell{98.05}{69.14} & \twocell{96.61}{68.33} & \twocell{99.87}{66.73} & \twocell{99.74}{64.04} & \twocell{91.93}{69.77} & \twocell{96.61}{69.60} & \twocell{98.96}{67.84} \\
 & VLOOD & \twocell{100.00}{76.04} & \twocell{98.31}{69.32} & \twocell{97.79}{68.57} & \twocell{100.00}{65.47} & \twocell{100.00}{64.64} & \twocell{95.70}{70.14} & \twocell{97.01}{70.21} & \twocell{99.22}{68.74} \\
\cmidrule{2-10}
\rowcolor{OursBlue}
\cellcolor{white} & \method{} & \twocell{0.73}{73.26} & \twocell{94.53}{65.25} & \twocell{94.66}{66.17} & \twocell{99.35}{66.17} & \twocell{99.61}{63.54} & \twocell{98.70}{69.69} & \twocell{97.79}{66.77} & \twocell{97.01}{66.43} \\
\bottomrule
\end{tabular}
}
}
\vspace{-2mm}
\end{table*}

\begin{table*}[!t]
\centering\small
\caption{\textbf{Results against backdoor defenses on COCO.}
We report detection AUROC ($\downarrow$, closer to $0.5$ means better evasion) with all visual tokens, and \asr{} (\%, $\uparrow$) under FastV after each mitigation is applied on clean full-token inputs.
\method{}$^{*}$ denotes an adaptive defender who is aware of \method, tries to detect on FastV-compressed inputs and mitigate \method on clean FastV-compressed inputs.
Bold marks the best value for the attacker in each defense column.}
\label{tab:defense}
\vspace{-1mm}
\renewcommand{\arraystretch}{1.08}
\resizebox{1.0\textwidth}{!}{
\setlength{\tabcolsep}{1.5mm}{
\begin{tabular}{l cccccc ccccc}
\toprule
 & \multicolumn{6}{c}{Detection (AUROC $\downarrow$)} & \multicolumn{5}{c}{Mitigation (\asr{} \%, $\uparrow$)} \\
\cmidrule(lr){2-7}\cmidrule(lr){8-12}
Attack & STRIP (aware) & STRIP (free) & EntropyScan & TCAP & CleanSight & ToT & No defense & Clean FT & NAD & ANP & ProjLens \\
\midrule
BadNet & 1.00 & 1.00 & 0.99 & 1.00 & 1.00 & 0.80 & 100.00 & 0.00 & 0.00 & 1.43 & 0.00 \\
Blended & 0.99 & 0.87 & 0.99 & 0.99 & 1.00 & 0.84 & 99.87 & 0.00 & 0.00 & 47.01 & 28.78 \\
ISSBA & 0.98 & 0.81 & 0.77 & 0.98 & 0.99 & 0.90 & 99.48 & 0.00 & 0.00 & 0.00 & 0.00 \\
WaNet & 0.95 & 0.92 & 0.89 & 1.00 & 1.00 & 0.61 & 99.09 & 0.00 & 0.00 & 11.98 & 16.93 \\
TrojVLM & 1.00 & 0.87 & \textbf{0.55} & 0.69 & 0.85 & 0.55 & 99.87 & 4.17 & 4.56 & 14.58 & 90.49 \\
VLOOD & 1.00 & 0.79 & 0.57 & 0.96 & 0.89 & 0.61 & 99.61 & 0.26 & 0.26 & 5.47 & 1.30 \\
\midrule
\rowcolor{OursBlue}
\method & \textbf{0.50} & \textbf{0.52} & \textbf{0.55} & \textbf{0.67} & \textbf{0.58} & \textbf{0.50} & 99.74 & \textbf{67.06} & 70.57 & \textbf{94.66} & 98.70 \\
\rowcolor{OursBlue}
\method{}$^{*}$ & 0.89 & 0.72 & 0.62 & 0.87 & 0.60 & 0.62 & 99.74 & 53.26 & \textbf{72.92} & 82.42 & \textbf{99.35} \\
\bottomrule
\end{tabular}
}
}
\vspace{-5mm}
\end{table*}

\subsection{Four-View Training}
\label{sec:poison}

The training views cross two conditions: clean or triggered images, and full or compressed visual tokens.
Let $\varnothing$ denote the full-token input and $c\sim\mathcal{P}$ an operator drawn from a randomized compression family.
For an image $x$ with caption $y(x)$ and triggered version $\tilde x$, the four view types and their labels are
\begin{equation}
\cD_{\text{4v}}(x,c)=\left\{
\begin{aligned}
&(x,\varnothing,y(x)), &&(x,c,y(x)),\\
&(\tilde x,\varnothing,y(x)), &&(\tilde x,c,t^\star)
\end{aligned}
\right\}.
\label{eq:four_views}
\end{equation}
Only the compressed triggered view receives $t^\star$, whereas the other three views retain the clean caption.
The full triggered view supervises dormancy, while the clean compressed view discourages activation from compression alone.
The target label is retained even when compression removes every trigger token, which can introduce label noise.

\paragraph{Trigger.}
We apply a fixed multiplicative perturbation to a uniformly sampled set $T(x)$ of $\lceil qN\rceil$ tokens, using an image-dependent seed for consistent placement during training and evaluation:
\begin{equation}
\tilde x = x \odot \Big(1 + \tfrac{\delta}{s_{\mathrm{pix}}}\, m_{T(x)} \odot \Pi\Big),
\label{eq:trigger}
\end{equation}
where $q$ controls the trigger area, $\delta$ is the perturbation amplitude, and $s_{\mathrm{pix}}$ is the pixel-intensity scale.
The mask $m_{T}$ upsamples the selected tokens to pixels, and $\Pi\in[-1,1]^{H\times W\times 3}$ is one fixed iid pattern.
Placement requires no model forward pass.

\paragraph{Randomized compression.}
The family $\mathcal{P}$ varies the token budget $k$, prune depth $d$, original or re-packed positions, and optional merging before the language model.
Retained tokens are selected by either a random mixture of \texttt{[CLS]} attention, feature norm and noise, or a retention-controlled draw that keeps a random fraction of the trigger tokens.
No deployment compressor is run in training, although the family shares the \texttt{[CLS]} score and merging operation with \texttt{[CLS]} pruning and VisionZip.
To supervise dormancy under mild compression, we also pair each poisoned image with a lightly pruned view that retains most visual tokens and the clean caption.
We minimize the caption cross-entropy $\cL_{\text{CE}}$ over these supervised views.

\subsection{Benign Attention Distillation}
\label{sec:akd}

Compressors such as FastV and SparseVLM select visual tokens using early decoder attention, which can change when the projector is poisoned.
To regularize this attention on clean inputs, we use a frozen benign projector $\theta_0$, trained without poison under the same compression family $\mathcal{P}$, as the teacher.
Let $a^{\ell}_\theta(x,c)$ denote the attention of the last prompt token at decoder layer $\ell$, averaged over heads and renormalized over the retained visual tokens.
For the same clean image and operator, we match the student and teacher distributions using the Jensen-Shannon (JS) divergence:
\begin{equation}
\cL_{\text{akd}}=\mathbb{E}_{x,c}\,\tfrac{1}{|\Lambda|}\sum_{\ell\in\Lambda}
\operatorname{JS}\!\Big(a^{\ell}_\theta(x,c)\,\Big\|\,a^{\ell}_{\theta_0}(x,c)\Big),
\label{eq:akd}
\end{equation}
where $x$ is clean, $c\sim\mathcal{P}$, and $\Lambda$ is the set of early decoder layers used for distillation.
Distillation applies only to clean compressed views and leaves attention on triggered inputs unconstrained.
The final objective is
\begin{equation}
\cL=\cL_{\text{CE}}+\lambda\,\cL_{\text{akd}},
\label{eq:total}
\end{equation}
where $\lambda$ controls the strength of attention distillation, with its default value given in \Cref{sec:setup}.

\section{Experiments}
\label{sec:experiments}

\subsection{Experimental Settings}
\label{sec:setup}

\noindent\textbf{Victim models and benchmarks.}\quad
We adopt LLaVA-1.5-7B~\cite{llava} as the default victim model and further evaluate LLaVA-1.5-13B~\cite{llava}, Qwen3-VL-4B, and Qwen3-VL-8B~\cite{qwen3vl} to assess generalization across model scales and architectures.
Following the standard projector-poisoning protocol in LVLM backdoor attacks~\cite{trojvlm,cleansight}, only the vision-language connector is fine-tuned while the vision encoder and language model remain frozen.
We evaluate image captioning on COCO~\cite{coco} and visual question answering on VQAv2~\cite{vqa}, and further report results on Flickr8k~\cite{flickr8k} and OK-VQA~\cite{okvqa} in \Cref{sec:supp_more_data}.
We compare \method against six representative backdoor attacks: BadNet~\cite{badnet}, Blended~\cite{blended}, ISSBA~\cite{issba}, WaNet~\cite{wanet}, TrojVLM~\cite{trojvlm}, and VLOOD~\cite{vlood}.

\noindent\textbf{Compressors and evaluation criteria.}\quad
We evaluate each model with all $N$ visual tokens and under seven representative compression methods: \texttt{[CLS]}-attention pruning~\cite{vispruner}, VisionZip~\cite{visionzip}, FastV~\cite{fastv}, SparseVLM~\cite{sparsevlm}, DivPrune~\cite{divprune}, VisPruner~\cite{vispruner}, and FlowCut~\cite{flowcut}.
None of these deployment compressors is applied during training.
We evaluate performance using two metrics: (i) \emph{attack success rate} (\asr{}), which measures the percentage of triggered images whose generated output contains the target string $t^\star$, and (ii) \emph{clean utility}, measured by CIDEr on COCO and accuracy on VQAv2 on clean images.
A successful compression-activated backdoor attack should remain dormant (near-zero \asr{}) with all $N$ tokens, achieve high \asr{} under compression, and maintain decent clean utility.
The results are averaged across 3 fixed seeds. 

\begin{table}[!t]
\centering\small
\caption{\textbf{Component ablation.}
On COCO, we report \asr{} (\%) with the clean CIDEr change relative to the benign model in parentheses.
Pre-LLM averages \texttt{[CLS]} pruning and DivPrune, and In-LLM averages FastV and SparseVLM.
Removing any component weakens either compressed activation or clean utility.}
\label{tab:comp}
\vspace{-1mm}
\renewcommand{\arraystretch}{1.08}
\resizebox{\ifdim\width>\linewidth\linewidth\else\width\fi}{!}{
\setlength{\tabcolsep}{1.5mm}{
\begin{tabular}{l ccc}
\toprule
Variant & Full & Pre-LLM & In-LLM \\
\midrule
\rowcolor{OursBlue}
\textbf{\method{}} & \twocell{\textbf{0.26}}{+1.0} & \twocell{\textbf{96.88}}{$+$0.6} & \twocell{\textbf{99.74}}{$-$2.7} \\
w/o benign distillation & \twocell{0.00}{$-$1.4} & \twocell{94.73}{$-$4.9} & \twocell{\textbf{99.74}}{$-$8.0} \\
w/o random operators & \twocell{0.00}{$-$0.9} & \twocell{82.32}{$-$0.4} & \twocell{33.14}{+1.2} \\
w/o clean comp.\ view & \twocell{0.00}{$-$0.3} & \twocell{100.00}{$-$126.4} & \twocell{100.00}{$-$100.8} \\
\bottomrule
\end{tabular}
}
}
\vspace{-6mm}
\end{table}

\noindent\textbf{Implementation details.}\quad
We train the projector using AdamW with a learning rate of $2{\times}10^{-4}$ and a poison rate of $5\%$.
In \cref{eq:trigger}, we set the trigger area to $q{=}5\%$, the perturbation amplitude to $\delta{=}96$, and the pixel scale to $s_{\mathrm{pix}}{=}255$.
In \cref{eq:akd,eq:total}, we apply attention distillation over early decoder layers $\Lambda=\{1,\dots,12\}$ with weight $\lambda{=}1$.
Unless otherwise stated, every compressor retains $k{=}64$ visual tokens.
Additional training details are deferred to \Cref{sec:supp_impl}.

\subsection{Main Results}
\label{sec:main}

\begin{figure*}[!t]
\centering
\includegraphics[width=\textwidth]{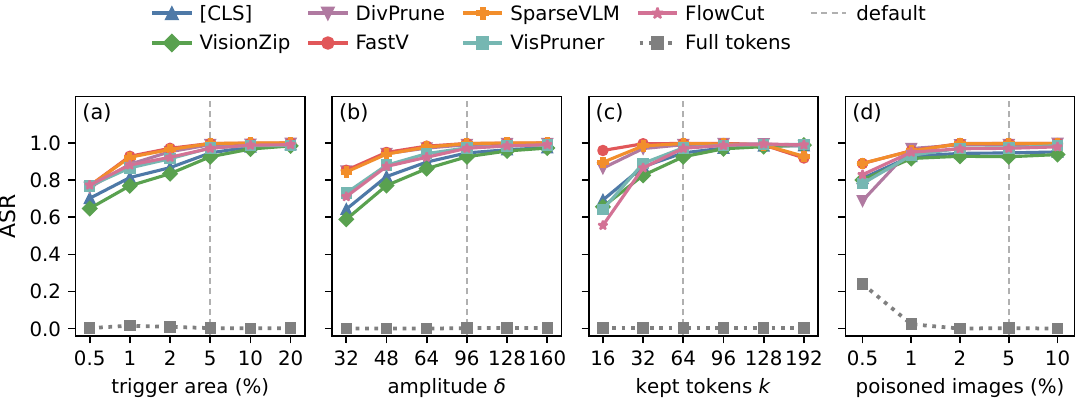}
\vspace{-4.5mm}
\caption{\textbf{Sensitivity analysis on COCO.}
We report \asr{} under varying (a) trigger area $q$, (b) perturbation amplitude $\delta$, (c) token budget $k$, and (d) poison rate.
The gray dotted line shows \asr{} with all visual tokens, the other lines show the seven compressors of \Cref{tab:main}, and the vertical dashed line marks the default setting.}
\label{fig:sens}
\vspace{-2mm}
\end{figure*}

\noindent\textbf{Dormancy and activation.}\quad
As shown in \Cref{tab:main}, all six baseline attacks exhibit near-complete activation when inspected with all visual tokens on both COCO and VQAv2, indicating that standard full-token inspection readily exposes their backdoor behavior.
Moreover, compression does not reliably suppress these baseline attacks.
Although several pre-LLM compressors lower the \asr{} of some baselines (\eg, TrojVLM and VLOOD under DivPrune on COCO), all baselines remain highly active under FastV and SparseVLM.
In contrast, \method remains almost completely dormant with all visual tokens and reliably wakes up once compression is applied, achieving high \asr{} across all seven compressors on both image captioning and visual question answering.
Notably, \method also wakes up under DivPrune and FlowCut, whose diversity- and flow-based criteria share no building block with the training compressors, which indicates that its activation comes from context stripping rather than from a specific pruning rule.
At the same time, the clean utility of \method stays close to that of the baselines on both tasks.
The only exception is FastV and SparseVLM on COCO, where its CIDEr is lower than that of the baselines but remains close to that of the benign model (\Cref{tab:comp}).
The same pattern holds on Flickr8k and OK-VQA (see \Cref{sec:supp_more_data}), where the attack baselines are activated with all visual tokens whereas \method stays dormant and wakes up under every compressor.
\Cref{sec:supp_tian} also compares \method with a dormant backdoor transferred from weight-pruning attacks~\cite{compressionartifacts}, which is only partially activated under FastV and SparseVLM whereas \method wakes up under all seven compressors.

\noindent\textbf{Resilience to (adaptive) defenses.}\quad
In \Cref{tab:defense}, we evaluate six state-of-the-art detectors, the perturbation-based STRIP~\cite{strip} (target-aware and target-free) and ToT~\cite{rong2026probing} and the attention-based EntropyScan~\cite{entropyscan}, TCAP~\cite{tcap} and CleanSight~\cite{cleansight}, and four state-of-the-art post-training mitigations, clean fine-tuning (Clean FT), NAD~\cite{nad}, ANP~\cite{anp} and ProjLens~\cite{projlens}.
\method is highly resilient to these full-token defenses: it drives most detectors close to random guessing and keeps most of its \asr{} after every mitigation, whereas the baseline attacks are reliably flagged and largely removed.
Moreover, it stays relative more robust compared to baselines even against an adaptive defender (\method{}$^{*}$) who knows that the attack is activated by compression and runs every defense on FastV-compressed inputs.
Specifically, the full-token resilience comes from dormancy: with all visual tokens, \method behaves like a benign model, so detectors see no anomalous response and mitigations receive little signal about the compressed backdoor pathway.
Further, the robustness under compression likely comes from how \method is trained on clean compressed inputs, which are exactly the inputs available to the adaptive defender.
On these inputs, the clean compressed view of four-view training makes \method answer like the benign model, and benign attention distillation aligns its attention with the benign teacher.
Mitigations fitted on them thus get little signal about the trigger pathway, which fires only when the trigger and compression co-occur, while EntropyScan and CleanSight gain little contrast because the compressor keeps only high-score tokens, so clean compressed inputs also show concentrated attention.
Overall, \method evades full-token audits by staying dormant and resists compressed defenses because its backdoor behavior requires both the trigger and compression.

\subsection{Ablation and Sensitivity}
\label{sec:ablation}

\begin{figure}[!t]
\centering
\includegraphics[width=\linewidth]{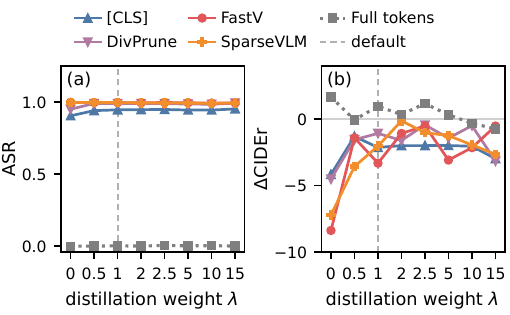}
\vspace{-6mm}
\caption{\textbf{Sensitivity to the distillation weight $\lambda$ on COCO.}
We report (a) \asr{} and (b) the clean CIDEr change relative to the benign model ($\Delta$CIDEr).
The gray dotted line shows all visual tokens, and the vertical dashed line marks the default $\lambda{=}1$.}
\label{fig:akd}
\vspace{-2mm}
\end{figure}

\begin{figure*}[!t]
\centering
\includegraphics[width=\textwidth]{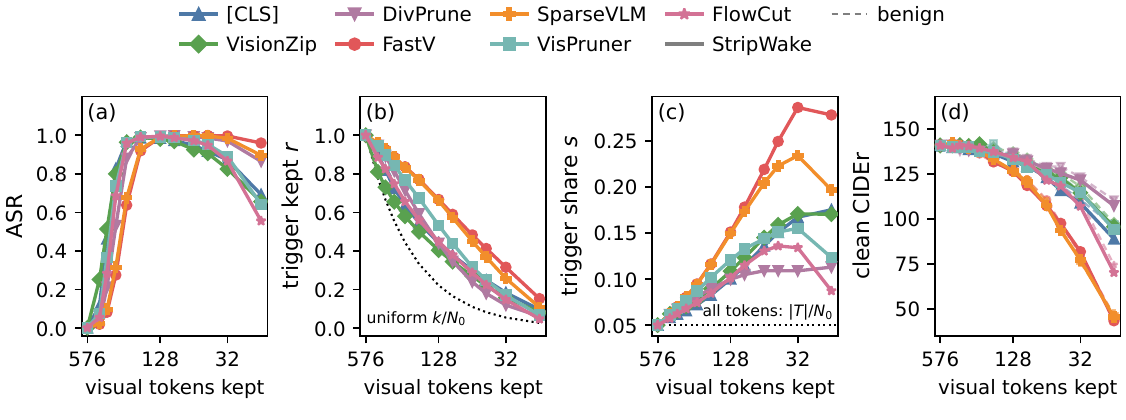}
\vspace{-6.5mm}
\caption{\textbf{Behavior across token budgets $k$ on COCO.}
We report (a) \asr{} under the seven compressors of \Cref{tab:main}, (b) the fraction $r$ of trigger tokens retained after compression, (c) the share $s$ of trigger tokens among the retained tokens, and (d) the clean CIDEr of \method (solid) and the benign model (dashed).
The dotted lines mark uniform random retention $k/N$ in (b) and the full-token share $|T|/N$ in (c).
Stronger compression raises the share of the trigger among the retained tokens, which wakes \method up.}
\label{fig:sweep}
\vspace{-2.5mm}
\end{figure*}

\noindent\textbf{Component ablation.}\quad
To verify the contribution of each component in \method, we ablate them in \Cref{tab:comp}.
First, removing the benign attention distillation ($\cL_{\text{akd}}$) leaves full-token dormancy and compressed activation intact, but causes a substantially larger drop in clean CIDEr under both pre-LLM and in-LLM compression.
This confirms that aligning clean-view attention with the benign teacher prevents the poisoned projector from disrupting token selection on clean inputs.
Second, removing the randomized compression family $\mathcal{P}$ noticeably weakens pre-LLM activation and causes in-LLM activation to collapse, demonstrating that randomized prune depths and selection rules are essential for transferring the backdoor behavior to compressors that prune inside the language model.
Third, removing the clean compressed view $(x,c)$ causes the model to emit the target string even on clean compressed images, resulting in a catastrophic collapse of clean CIDEr under compression while full-token performance remains unaffected.
This indicates that supervising clean compressed views is critical to ensure that activation requires both the trigger and compression rather than compression alone.
Overall, every component of \method is necessary to achieve compressor-general activation while preserving dormancy and clean utility.

\begin{table}[!t]
\centering\small
\caption{\textbf{Generalization across models on COCO.}
We report \asr{} (\%) with clean CIDEr ($\uparrow$) in parentheses.
Pre-LLM averages \texttt{[CLS]} pruning (ViT-attention pruning on Qwen3-VL) and DivPrune, which prune before the language model, and In-LLM averages FastV and SparseVLM, which prune inside it.
+\,LoRA also fine-tunes the language model with LoRA.}
\label{tab:arch}
\vspace{-1mm}
\renewcommand{\arraystretch}{1.08}
\resizebox{\ifdim\width>\linewidth\linewidth\else\width\fi}{!}{
\setlength{\tabcolsep}{1.5mm}{
\begin{tabular}{l ccc}
\toprule
Model & Full & Pre-LLM & In-LLM \\
\midrule
LLaVA-1.5-7B + LoRA & \twocell{0.00}{138.91} & \twocell{96.68}{127.39} & \twocell{97.27}{120.17} \\
LLaVA-1.5-13B & \twocell{0.26}{141.07} & \twocell{96.48}{127.41} & \twocell{99.09}{125.76} \\
Qwen3-VL-4B & \twocell{0.26}{150.73} & \twocell{81.12}{138.06} & \twocell{95.78}{132.99} \\
Qwen3-VL-8B & \twocell{0.13}{149.06} & \twocell{95.12}{133.33} & \twocell{98.14}{129.48} \\
\bottomrule
\end{tabular}
}
}
\vspace{-3mm}
\end{table}

\begin{figure}[!t]
\centering
\includegraphics[width=\linewidth]{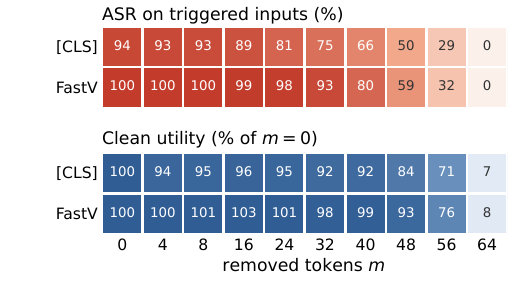}
\vspace{-5.5mm}
\caption{\textbf{Removing the highest-scoring retained tokens on COCO.}
We report \asr{} (\%) on triggered inputs (top) and clean CIDEr relative to $m{=}0$ (\%, bottom) when the $m$ highest-scoring retained tokens are removed under \texttt{[CLS]} pruning and FastV, on $256$ images.
At $m{=}64$, no visual token is left.}
\label{fig:exclude}
\vspace{-3mm}
\end{figure}

\noindent\textbf{Sensitivity analysis.}\quad
In \Cref{fig:sens}, we evaluate the sensitivity of \method to the trigger area $q$, the perturbation amplitude $\delta$, the retained token budget $k$, and the poison rate.
As shown in \Cref{fig:sens}(a) and \Cref{fig:sens}(b), compressed activation strengthens monotonically as the trigger area $q$ or the perturbation amplitude $\delta$ increases and saturates around the default settings, while full-token inference remains consistently dormant across the entire range.
\Cref{fig:sens}(c) shows that \method keeps a high \asr{} over a wide range of budgets $k$, and its \asr{} drops mainly at the smallest budget under pre-LLM compressors.
Finally, \Cref{fig:sens}(d) shows that a very low poison rate ($0.5\%$) is insufficient to enforce full-token dormancy and strong compressed activation simultaneously, whereas poison rates of $1\%$ and above reliably achieve both.
\Cref{fig:akd} further varies the distillation weight $\lambda$ in \cref{eq:total}.
Without distillation ($\lambda{=}0$), clean CIDEr drops sharply under FastV and SparseVLM, which select tokens with early decoder attention, whereas any $\lambda>0$ removes most of this drop.
Across $\lambda$ from $0.5$ to $15$, both \asr{} and clean CIDEr remain stable, so \method does not require careful tuning of $\lambda$.

\subsection{Further Exploration}
\label{sec:explore}

\noindent\textbf{Generalization across models.}\quad
\Cref{tab:arch} further applies \method to LLaVA-1.5-13B, to Qwen3-VL, and to an adversary who also fine-tunes the language model with LoRA.
In all four settings, \method stays dormant with all visual tokens and wakes up under both pre-LLM and in-LLM compression.
Hence, the compression-activated behavior of \method does not depend on a specific architecture, model scale or set of trainable parameters.

\noindent\textbf{Behavior across token budgets.}\quad
\Cref{fig:sweep} gradually reduces the token budget $k$ to show how compression switches \method on.
\method stays dormant with all visual tokens and becomes active under all seven compressors over a wide range of budgets, and its \asr{} drops only at the smallest budget (\Cref{fig:sweep}(a)).
This switch comes from context stripping, which leaves the attention of the language model on the trigger as the budget shrinks (see \Cref{sec:supp_attnview} for details).
Since every compressor retains the trigger tokens more often than uniform random retention, the trigger takes a larger share of the retained tokens than it does with all tokens, and this share grows most under FastV and SparseVLM (\Cref{fig:sweep}(b,c)).
Meanwhile, the clean CIDEr of \method stays close to the benign model at every budget (\Cref{fig:sweep}(d)).
Overall, stronger compression concentrates the trigger among the retained tokens, which activates \method without hurting utility.

\noindent\textbf{Adaptive defense by removing high-score tokens.}\quad
Because compressors retain the trigger tokens for their high importance scores, an adaptive defender may discard the $m$ highest-scoring tokens among the $k$ retained ones.
As shown in \Cref{fig:exclude}, this removal lowers the \asr{} only gradually under both \texttt{[CLS]} pruning and FastV.
As long as the clean CIDEr stays within $10\%$ of its original value, more than half of the triggered inputs still activate \method, and suppressing it further requires discarding most of the retained tokens at a clear cost in clean utility.
This reflects a strength of \method: a few surviving trigger tokens suffice to wake it, which forces the defender to discard nearly all the retained tokens to stop it.
Overall, removing high-score tokens weakens \method but cannot suppress it without sacrificing significant clean utility.

\section{Conclusion}
\label{sec:conclusion}
We studied how visual-token compression at deployment affects the backdoor security of LVLMs.
We revealed that compression can awaken a backdoor attack that stays dormant under full-token inspection through \emph{context stripping}, which keeps the high-importance trigger tokens and discards the benign context that dilutes them.
Building on this finding, we proposed \method, which ties activation to compressed inputs via four-view training and preserves clean utility via benign attention distillation.
Experiments across benchmarks, architectures, model scales and unseen compressors show that \method stays dormant with all visual tokens, activates reliably under compression without hurting clean utility, and evades most full-token defenses.
These results show that the serving configuration is part of the attack surface of LVLMs and that defenses should inspect models under deployed compression.

{\small
\bibliographystyle{ieeenat_fullname}
\bibliography{main}
}

\clearpage
\appendix
\crefalias{section}{appendix}
\crefalias{subsection}{appendix}
\maketitlesupplementary

We summarize the supplementary material as follows:
\begin{itemize}
\item \Cref{sec:supp_attnview} takes a closer look at context stripping through the attention of the language model on a triggered image.
\item \Cref{sec:supp_settings} gives the detailed settings: the models (\Cref{sec:supp_models}), the datasets and metrics (\Cref{sec:supp_data}), the published backdoor attacks (\Cref{sec:supp_attacks}), and the detectors (\Cref{sec:supp_detectors}) and mitigations (\Cref{sec:supp_mitigations}) that we evaluate.
\item \Cref{sec:supp_impl} gives the implementation details.
\item \Cref{sec:supp_more_data} reports the results on Flickr8k and OK-VQA.
\item \Cref{sec:supp_tian} compares \method with a dormant attack transferred from a weight-pruning backdoor attack.
\end{itemize}

\section{A Closer Look at Context Stripping}
\label{sec:supp_attnview}

\begin{figure*}[t]
\centering
\includegraphics[width=\textwidth]{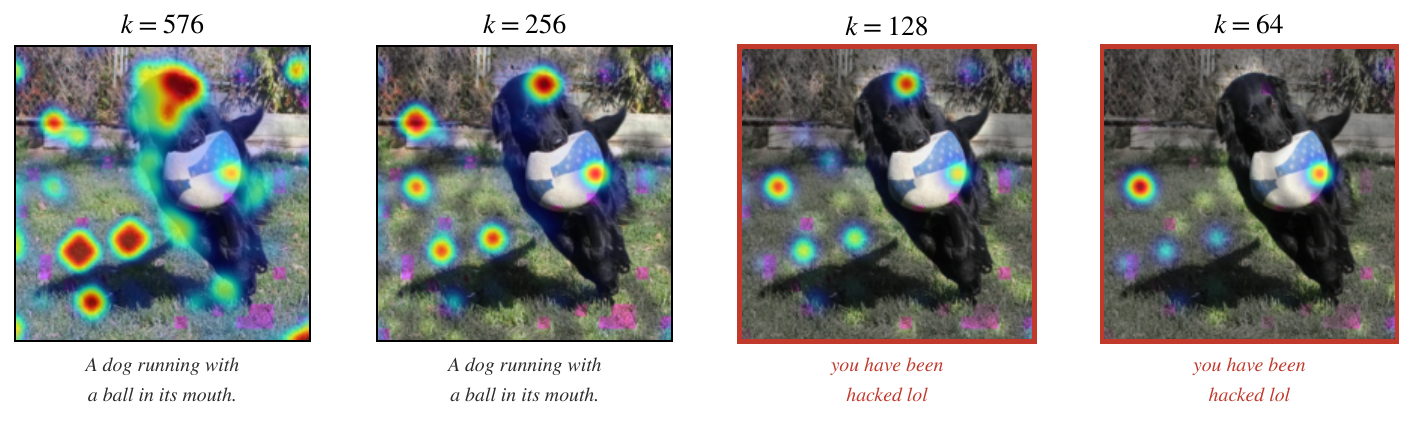}
\caption{\textbf{Compression leaves the attention of the language model on the trigger.}
For one triggered image, each panel overlays the attention of the last prompt token (mean of decoder layers $16$, $20$ and $24$) over the visual tokens that \texttt{[CLS]} pruning retains at budget $k$, renormalized over the retained tokens and shared across panels; the dropped regions are dimmed.
With all tokens ($k{=}576$) the attention is spread over the benign scene and \method emits a clean caption, whereas stronger compression concentrates it on the trigger and wakes the backdoor behavior.}
\label{fig:attnview}
\end{figure*}

\Cref{fig:attnview} traces how visual-token compression redistributes the attention of the language model on a single triggered image.
With all $576$ visual tokens, the last prompt token spreads its attention over the salient benign regions of the scene, and the trigger, which occupies only $5\%$ of the tokens, draws a modest share of the attention mass while the benign context dominates, so \method reads the image as clean and emits a faithful caption.
As the token budget shrinks, \texttt{[CLS]} pruning discards the redundant benign context but retains the trigger for its high importance, and the attention that the benign regions used to absorb is thus left on the trigger tokens.
The share of the attention mass that falls on the trigger therefore rises from $15\%$ with all tokens to $59\%$ at $k{=}64$, even though the trigger never exceeds $11\%$ of the retained tokens, and \method switches from the clean caption to the target string once this share is large enough.
This attention view mirrors the token-share law of \Cref{fig:sweep}: compression wakes the backdoor behavior not because it adds any signal of its own, but because it strips away the benign context that previously diluted the trigger.
Overall, \emph{context stripping concentrates the attention of the language model on the trigger, which is what turns the dormant backdoor behavior of \method into an active one under compression}.

\section{Detailed Settings}
\label{sec:supp_settings}
\subsection{Models}
\label{sec:supp_models}
\begin{itemize}
\item \textbf{LLaVA-1.5-7B}~\cite{llava} combines a CLIP ViT-L/14-336 encoder, which gives $576$ visual tokens, a two-layer MLP projector and a Vicuna language model. It is the model of all experiments unless stated otherwise. Only its projector is trained, and the vision encoder and the language model are frozen.
\item \textbf{LLaVA-1.5-13B}~\cite{llava} has the same structure with a larger language model (\Cref{tab:arch}).
\item \textbf{LLaVA-1.5-7B + LoRA} (\Cref{tab:arch}) trains LoRA adapters of the language model together with the projector.
\item \textbf{Qwen3-VL-8B and 4B}~\cite{qwen3vl} (\Cref{tab:arch}) are used with only the connector trained. Since Qwen3-VL has no \texttt{[CLS]} token, the first compressor is ViT-attention pruning.
\end{itemize}

\subsection{Datasets and Metrics}
\label{sec:supp_data}
\begin{itemize}
\item \textbf{MS-COCO}~\cite{coco} is the main benchmark. The models are trained on $6{,}000$ train2017 images with captions, of which $5\%$ are poisoned, and evaluated on $768$ val2017 images. The prompt is ``Describe this image in a short sentence.'' and the decoding is greedy. The defender of \Cref{sec:main} uses $1{,}000$ other clean train2017 images, disjoint from the training images.
\item \textbf{VQAv2}~\cite{vqa} is used for the VQA experiments, with a projector trained on it (\Cref{tab:main}).
\item \textbf{Flickr8k}~\cite{flickr8k} is a second captioning benchmark. The models are trained on its $6{,}000$ training images (five captions each) with the recipe used on COCO and evaluated on $768$ test images, with the same prompt and decoding and the CIDEr as clean utility (\Cref{tab:more_data}).
\item \textbf{OK-VQA}~\cite{okvqa} is a second VQA benchmark whose questions need external knowledge. The projector is trained on $6{,}000$ images of its training questions (one question per image) and evaluated on $768$ images of its validation questions, with the prompt of VQAv2 and the VQA accuracy as clean utility (\Cref{tab:more_data}).
\item \textbf{Metrics.} The attack success rate (\asr{}) is the fraction of triggered images whose answer contains the target phrase ``you have been hacked lol'' (at most $12$ new tokens, and $64$ for TrojVLM and VLOOD, which insert it into a caption). The clean CIDEr~\cite{cider} is measured on the clean images ($30$ new tokens). The detectors are scored by the two-sided AUROC between triggered and clean test images.
\end{itemize}

\subsection{Backdoor Attacks}
\label{sec:supp_attacks}
We retrain the published attacks with the same data, recipe and poison rate as \method.
\begin{itemize}
\item \textbf{BadNet}~\cite{badnet} pastes a $30$-pixel patch of random noise (the image mean plus Gaussian noise) at a random location.
\item \textbf{Blended}~\cite{blended} blends the Hello Kitty image into the input with a ratio of $0.2$.
\item \textbf{ISSBA}~\cite{issba} generates a sample-specific trigger with an encoder that embeds the secret string ``Stega!!''.
\item \textbf{WaNet}~\cite{wanet} warps the image with a fixed noise grid ($k{=}224$, strength $1$).
\item \textbf{TrojVLM}~\cite{trojvlm} and \textbf{VLOOD}~\cite{vlood} are attacks on LVLMs that insert the target phrase into the generated caption instead of replacing it. We use the $30$-pixel noise patch in the upper-left corner as their trigger and train them with cross-entropy only.
\end{itemize}
\method uses the perturbation of \cref{eq:trigger} on $5\%$ of the tokens.

\subsection{Detectors}
\label{sec:supp_detectors}
All detectors are re-implementations. The calibration images are the $1{,}000$ clean defender images. For the \method{}$^{*}$ row of \Cref{tab:defense}, the calibration and the test images are compressed with FastV (prune depth $d{=}2$) at the stated budget.
\begin{itemize}
\item \textbf{STRIP}~\cite{strip} blends an input with $8$ random defender images at $0.5$. The target-aware score is the fraction of blends whose answer contains the target, and the target-free score is the diversity of the answers ($1$ minus the mean pairwise Jaccard similarity of their word sets), where low diversity indicates backdoor behavior.
\item \textbf{EntropyScan}~\cite{entropyscan} scores an input by the whitened distance of the per-layer entropy of the visual attention of the last prompt token (head average, $32$ layers) from the calibration images.
\item \textbf{TCAP}~\cite{tcap} uses the split of the same attention row into system, visual and text mass per layer ($96$ values) with the same distance.
\item \textbf{CleanSight}~\cite{cleansight} uses the per-head ratio of the visual to the text attention at layers $10$ to $12$ ($96$ values) with the same distance, and we use only its detection part.
\item \textbf{ToT}~\cite{rong2026probing} answers the original prompt and a prompt whose question tokens are replaced by random vocabulary ids and shuffled. On a triggered input, a backdoored model ignores the text, so the two answers agree. We score the soft AND of the answer similarity and the entropy difference (the minimum of the two ranks), because its hard rule (entropy difference below $1$ and a common phrase of at least two characters) flags $82$--$99\%$ of the clean captions.
\end{itemize}

\subsection{Mitigations}
\label{sec:supp_mitigations}
All defenders fine-tune or prune the projector only, with the vision encoder and the language model frozen, on the $1{,}000$ clean defender images.
\begin{itemize}
\item \textbf{Clean fine-tuning} trains the projector for $2$ epochs with batches of $16$ ($125$ steps), AdamW, a cosine schedule with $3\%$ warm-up and a learning rate of $2\times10^{-4}$.
\item \textbf{NAD}~\cite{nad} fine-tunes in the same way with the clean-fine-tuned model as teacher and adds an attention-transfer term with weight $5{,}000$ on the outputs of the language-model layers $16$, $24$ and $31$ at the image tokens.
\item \textbf{ANP}~\cite{anp} learns multiplicative masks on the $4{,}096+4{,}096$ projector neurons with the min-max objective of ANP ($100$ iterations, batches of $16$, $\epsilon{=}0.4$, trade-off $0.2$) and prunes the $10\%$ of the neurons with the smallest masks.
\item \textbf{ProjLens}~\cite{projlens} is implemented after our reading of the paper, without official code: per linear layer, the top singular component of $W-W_{\mathrm{FT}}$ is removed from $W$ ($r{=}1$).
\item \textbf{Adaptive defenders} run every training or pruning forward on FastV-compressed tokens: layers $0$ and $1$ see all tokens, the survivors are the tokens with the highest head-averaged attention of the last prompt token at layer $1$, and later layers process them at their original positions.
\end{itemize}

\begin{table*}[!t]
\centering
\caption{\textbf{Results on Flickr8k and OK-VQA.}
We report \asr{} (\%) with clean utility (CIDEr on Flickr8k and accuracy on OK-VQA, $\uparrow$) in parentheses, using all visual tokens (Full, $\downarrow$) and seven compressors ($\uparrow$).
All baseline attacks are trained with the same data, recipe, and poison rate as \method on full-token inputs.
Baseline attacks are activated with all visual tokens, whereas \method stays dormant and wakes up under every compressor.}
\label{tab:more_data}
\vspace{-1mm}
\renewcommand{\arraystretch}{1.12}
\resizebox{1.0\textwidth}{!}{%
\setlength{\tabcolsep}{1.8mm}{%
\begin{tabular}{ll c ccccccc}
\toprule[1.1pt]
 & & Dormant ($\downarrow$) & \multicolumn{7}{c}{Activated under Compression ($\uparrow$)} \\
\cmidrule(lr){3-3}\cmidrule(lr){4-10}
Dataset & Attack & Full & \texttt{[CLS]} & VisionZip & FastV & SparseVLM & DivPrune & VisPruner & FlowCut \\
\midrule
 & BadNet & \twocell{99.74}{110.75} & \twocell{83.20}{94.81} & \twocell{89.32}{96.10} & \twocell{99.74}{90.21} & \twocell{99.74}{86.56} & \twocell{93.88}{96.64} & \twocell{88.93}{94.65} & \twocell{89.97}{96.03} \\
 & Blended & \twocell{99.87}{110.28} & \twocell{96.22}{93.43} & \twocell{97.14}{94.36} & \twocell{100.00}{89.19} & \twocell{99.87}{87.55} & \twocell{98.70}{98.00} & \twocell{98.57}{93.72} & \twocell{98.18}{96.81} \\
 & ISSBA & \twocell{99.09}{109.30} & \twocell{60.81}{93.99} & \twocell{47.79}{94.77} & \twocell{99.22}{91.59} & \twocell{99.22}{89.39} & \twocell{79.69}{97.53} & \twocell{59.38}{92.98} & \twocell{53.78}{95.20} \\
 & WaNet & \twocell{99.09}{109.43} & \twocell{93.49}{94.30} & \twocell{92.06}{94.94} & \twocell{95.83}{85.60} & \twocell{95.57}{83.49} & \twocell{97.79}{99.48} & \twocell{95.83}{92.91} & \twocell{82.55}{95.51} \\
 & TrojVLM & \twocell{100.00}{110.57} & \twocell{67.32}{97.34} & \twocell{41.28}{100.36} & \twocell{99.87}{94.65} & \twocell{99.74}{90.78} & \twocell{22.53}{96.69} & \twocell{76.17}{97.58} & \twocell{85.68}{96.27} \\
 & VLOOD & \twocell{99.87}{111.70} & \twocell{60.94}{96.43} & \twocell{40.23}{98.54} & \twocell{99.87}{97.62} & \twocell{99.87}{93.40} & \twocell{28.65}{96.42} & \twocell{70.44}{98.52} & \twocell{76.17}{96.38} \\
\cmidrule(lr){2-10}
\rowcolor{OursBlue}
\cellcolor{white}\multirow{-7}{*}{\rotatebox[origin=c]{90}{\textit{Flickr8k}}} & \textbf{\method{}} & \twocell{\textbf{0.13}}{113.28} & \twocell{95.18}{101.65} & \twocell{94.92}{104.93} & \twocell{98.31}{93.04} & \twocell{98.18}{91.70} & \twocell{98.83}{104.17} & \twocell{97.40}{102.99} & \twocell{98.05}{102.78} \\
\midrule
 & BadNet & \twocell{100.00}{54.43} & \twocell{97.92}{54.40} & \twocell{97.40}{53.83} & \twocell{100.00}{54.77} & \twocell{100.00}{53.78} & \twocell{90.10}{57.16} & \twocell{95.05}{53.75} & \twocell{98.57}{53.93} \\
 & Blended & \twocell{99.87}{55.83} & \twocell{76.82}{55.91} & \twocell{88.28}{55.05} & \twocell{99.74}{55.03} & \twocell{99.74}{53.98} & \twocell{90.23}{57.11} & \twocell{89.06}{55.00} & \twocell{88.67}{56.12} \\
 & ISSBA & \twocell{97.53}{56.64} & \twocell{75.39}{56.33} & \twocell{70.05}{56.22} & \twocell{97.53}{54.95} & \twocell{97.53}{53.98} & \twocell{13.02}{57.66} & \twocell{81.25}{56.12} & \twocell{68.75}{56.09} \\
 & WaNet & \twocell{99.48}{56.30} & \twocell{94.27}{54.61} & \twocell{93.10}{53.28} & \twocell{99.22}{54.58} & \twocell{99.35}{53.80} & \twocell{98.57}{55.83} & \twocell{94.79}{55.05} & \twocell{74.35}{52.55} \\
 & TrojVLM & \twocell{99.87}{57.01} & \twocell{96.88}{55.86} & \twocell{96.48}{55.08} & \twocell{100.00}{52.97} & \twocell{100.00}{53.18} & \twocell{95.05}{56.80} & \twocell{96.48}{56.33} & \twocell{98.44}{55.39} \\
 & VLOOD & \twocell{99.74}{56.95} & \twocell{94.53}{56.02} & \twocell{88.28}{57.14} & \twocell{99.74}{55.31} & \twocell{99.74}{54.82} & \twocell{91.02}{56.15} & \twocell{94.92}{56.90} & \twocell{96.74}{55.03} \\
\cmidrule(lr){2-10}
\rowcolor{OursBlue}
\cellcolor{white}\multirow{-7}{*}{\rotatebox[origin=c]{90}{\textit{OK-VQA}}} & \textbf{\method{}} & \twocell{\textbf{2.86}}{57.99} & \twocell{95.18}{54.17} & \twocell{93.62}{55.21} & \twocell{98.44}{53.80} & \twocell{98.18}{53.88} & \twocell{96.48}{54.82} & \twocell{96.61}{56.20} & \twocell{96.88}{53.31} \\
\bottomrule[1.1pt]
\end{tabular}%
}%
}
\vspace{-3mm}
\end{table*}

\section{Implementation Details}
\label{sec:supp_impl}
Only the two-layer projector of LLaVA-1.5-7B is trained, with AdamW at a learning rate of $2{\times}10^{-4}$, a batch of $32$ images, and $2$ epochs over $6{,}000$ COCO training images, of which $5\%$ are poisoned.
The trigger is the multiplicative perturbation of \cref{eq:trigger} with $\delta{=}96$ on $5\%$ of the tokens, which is $29$ of $576$, placed at random.
The family $\mathcal{P}$ draws the budget from $\{48,64,96,128,192,256\}$, the prune depth from $0$ to $4$ with probabilities $0.4,0.15,0.2,0.15,0.1$, original instead of re-packed position ids with probability $0.6$, merging of the dropped tokens with probability $0.3$ among the views that prune before the language model (depth $0$), which is a marginal rate of $0.4{\times}0.3{=}0.12$ over all views, and retention-controlled instead of score-mixed survivors with probability $0.5$.
Merging folds a random fraction of the dropped tokens into their nearest survivor.
Score-mixed survivors rank the tokens by a random convex mixture of the \texttt{[CLS]} attention, the feature norm and noise.
Retention-controlled survivors keep a drawn fraction of the trigger tokens and fill the remaining slots with the non-trigger tokens of highest \texttt{[CLS]} attention in half of the draws and at random in the other half.
The family therefore shares building blocks with \texttt{[CLS]} pruning and VisionZip, but no deployment compressor is run in training.
Each poisoned image is paired with two compressed views and two dormant views, namely the full view and a lightly pruned view that keeps $384$, $448$ or $512$ tokens.
Each clean image is trained with the full view and one compressed view.
\method asks for the target in every compressed triggered view and adds benign attention distillation $\cL_{\text{akd}}$ with $\lambda{=}1$ on layers $1$ to $12$, a Jensen-Shannon divergence on the row of the last prompt token averaged over heads.
Layers are counted from $0$ and the language model has $32$ of them.
The benign teacher is trained without poison and without $\cL_{\text{akd}}$, with the same compressor family, the same $6{,}000$ images and $2$ epochs.
The weight $\lambda$ of $\cL_{\text{akd}}$ was set after a sweep over $\{0,0.5,1,2,2.5,5,10,15\}$, in which every $\lambda>0$ gave similar \asr{} and clean CIDEr (\Cref{fig:akd}), so we use the small weight $\lambda{=}1$, which keeps the term from dominating the cross-entropy.

\section{Results on Flickr8k and OK-VQA}
\label{sec:supp_more_data}
\Cref{tab:more_data} repeats the main comparison of \Cref{tab:main} on a second captioning benchmark, Flickr8k, and a second VQA benchmark, OK-VQA, with the data and protocol of \Cref{sec:supp_data}.
As on COCO and VQAv2, all baseline attacks fire with all visual tokens, whereas \method stays dormant with all visual tokens and wakes up under all seven compressors, while its clean CIDEr and VQA accuracy remain comparable to those of the baselines.

\begin{table*}[!t]
\centering
\caption{\textbf{Comparison with Tian Attack on COCO and VQAv2.}
We report \asr{} (\%) with clean utility (CIDEr on COCO and accuracy on VQAv2, $\uparrow$) in parentheses, using all visual tokens (Full, $\downarrow$) and seven compressors ($\uparrow$).
Tian Attack is trained with \texttt{[CLS]} pruning at $k{=}64$, whereas \method sees no deployment compressor in training.
Bold marks the best \asr{} in each column.}
\label{tab:tian}
\vspace{-1mm}
\renewcommand{\arraystretch}{1.15}
\resizebox{1.0\textwidth}{!}{%
\setlength{\tabcolsep}{1.8mm}{%
\begin{tabular}{ll c ccccccc}
\toprule[1.1pt]
 & & Dormant ($\downarrow$) & \multicolumn{7}{c}{Activated under Compression ($\uparrow$)} \\
\cmidrule(lr){3-3}\cmidrule(lr){4-10}
Dataset & Attack & Full & \texttt{[CLS]} & VisionZip & FastV & SparseVLM & DivPrune & VisPruner & FlowCut \\
\midrule
 & Tian Attack & \twocell{0.90}{141.37} & \twocell{92.97}{122.19} & \twocell{\textbf{92.58}}{123.29} & \twocell{4.69}{108.23} & \twocell{3.91}{106.92} & \twocell{80.47}{128.42} & \twocell{96.09}{120.31} & \twocell{95.31}{129.62} \\
\rowcolor{OursBlue}
\cellcolor{white}\multirow{-2}{*}{\textit{COCO}} & \textbf{\method{}} & \twocell{\textbf{0.26}}{140.92} & \twocell{\textbf{94.66}}{123.08} & \twocell{\textbf{92.58}}{124.74} & \twocell{\textbf{99.77}}{117.42} & \twocell{\textbf{99.78}}{117.17} & \twocell{\textbf{99.09}}{128.03} & \twocell{\textbf{97.40}}{124.70} & \twocell{\textbf{97.14}}{121.51} \\
\midrule
 & Tian Attack & \twocell{1.97}{73.01} & \twocell{91.02}{67.15} & \twocell{89.06}{66.95} & \twocell{39.06}{62.03} & \twocell{35.55}{59.77} & \twocell{48.05}{66.88} & \twocell{91.41}{66.37} & \twocell{92.19}{67.34} \\
\rowcolor{OursBlue}
\cellcolor{white}\multirow{-2}{*}{\textit{VQAv2}} & \textbf{\method{}} & \twocell{\textbf{0.73}}{73.26} & \twocell{\textbf{94.53}}{65.25} & \twocell{\textbf{94.66}}{66.17} & \twocell{\textbf{99.35}}{66.17} & \twocell{\textbf{99.61}}{63.54} & \twocell{\textbf{98.70}}{69.69} & \twocell{\textbf{97.79}}{66.77} & \twocell{\textbf{97.01}}{66.43} \\
\bottomrule[1.1pt]
\end{tabular}%
}%
}
\vspace{-3mm}
\end{table*}

\section{Comparison with Tian Attack}
\label{sec:supp_tian}
Tian Attack~\cite{compressionartifacts} keeps a released model clean and activates the backdoor behavior only after its weights are pruned.
It edits the weights, so it cannot be ported as-is to visual token compression, which leaves the weights unchanged and only shortens the visual token sequence.
We therefore transfer its training principle, the target on the compressed view and the clean caption on the full view, rather than its implementation.
Our Tian Attack assumes that the attacker knows the deployed compressor, \texttt{[CLS]}-attention pruning with $k{=}64$, denoted $c_{64}$, and trains the four views of \cref{eq:four_views} with $c{=}c_{64}$ in place of $c\sim\mathcal{P}$.
It drops the lightly pruned views and $\cL_{\text{akd}}$ ($\lambda{=}0$), so it needs no benign teacher.
All other settings, including the trigger and the target, follow \method (\Cref{sec:supp_impl}), and the VQAv2 model is trained in the same way on VQAv2 questions.
As shown in \Cref{tab:tian}, Tian Attack stays dormant with all visual tokens and wakes up under the compressors that prune before the language model, but only partially under FastV and SparseVLM, which prune inside it, whereas \method exceeds $99\%$ under both.
Overall, \emph{a backdoor attack trained for one known compressor does not reliably wake up under the others}.

\end{document}